# Semiconductor SEM Image Defect Classification Using Supervised and Semi-Supervised Learning with Vision Transformers

(Topic/category: Defect Inspection and Reduction)

Chien-Fu (Frank) Huang[1], Katherine Sieg[1], Leonid Karlinksy[2], Nash Flores[1], Rebekah Sheraw[1], Xin Zhang[3]

[1]IBM Research, Albany, NY, [2]IBM Research, Cambridge MA, [3]IBM Research, Yorktown Heights, NY

chien-fu.frank.huang@ibm.com, xzhang@us.ibm.com

**Abstract:** Controlling defects in semiconductor processes is important for maintaining yield, improving production cost, and preventing time-dependent critical component failures. Electron beam-based imaging has been used as a tool to survey wafers in the line and inspect for defects. However, manual classification of images for these nano-scale defects is limited by time, labor constraints, and human biases. In recent years, deep learning computer vision algorithms have shown to be effective solutions for image-based inspection applications in industry. This work proposes application of vision transformer (ViT) neural networks for automatic defect classification (ADC) of scanning electron microscope (SEM) images of wafer defects. We evaluated our proposed methods on 300mm wafer semiconductor defect data from our fab in IBM Albany. We studied 11 defect types from over 7400 total images and investigated the potential of transfer learning of DinoV2 and semi-supervised learning for improved classification accuracy and efficient computation. We were able to achieve classification accuracies of over 90% with less than 15 images per defect class. Our work demonstrates the potential to apply the proposed framework for a platform agnostic in-house classification tool with faster turnaround time and flexibility.

## I. Introduction

In the semiconductor industry, it is critical to maintain a certain level of defectivity in wafers to ensure yield in product wafers. Defectivity level is monitored through regular inspections that require significant amounts of time and effort. Having various shapes and sizes, defects in semiconductors can be located anywhere in patterns, lead to critical failures, and limit yield. To improve yield and scale down human labor costs, semiconductor manufacturers have implemented automatic defect classification systems that scan the wafer surface and classification of defects from high magnification scanning electron microscope images [1]. Current ADC systems are limited in defect classification, necessitating human feedback which is limited due to bias. Therefore, research in development of more complex ADC systems implementing computer vision is required to improve classification accuracy.

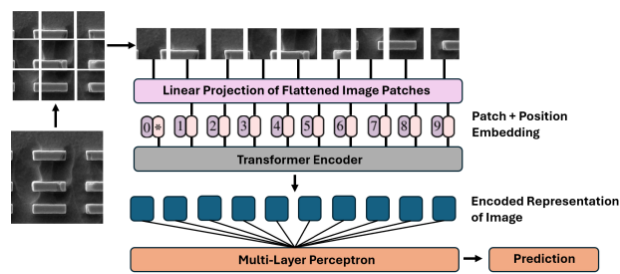

Fig. 1. Example of an SEM defect image being processed by a vision transformer architecture.

## II. Background Research Review

In the domain of computer vision, convolutional neural networks (CNN) have been the dominant model used. However, vision transformers have emerged as a noteworthy model for computer vision tasks. In 2017, transformers were introduced for natural language processing and later in 2020 they were adapted for computer vision tasks [2]. Since then, active work in industry and academia have been ongoing. While ViTs are more computationally expensive than CNNs, they scale better with data and are particularly useful when pre-trained on a large scale of data through self-supervised training.

The inner workings of a ViT model are shown in Figure 1. A ViT splits an image into patches, flattens the pixels in those patches, embeds those patches into vectors of dimension D (D depends on the model size), adds positional embeddings to retain positional information of the patches in the image, feeds it as input to a transformer encoder, and then outputs an encoded representation of the image. The encoded output can then be fed into a multi-layer perceptron (a type of neural network) to generate classification predictions for the image. This work studies the application of the prominent ViT model, DinoV2 [3]. DinoV2 is a set of self-supervised ViT models pretrained on large, curated data that have high performance applications in a range of computer vision tasks, namely image classification. The curated data that these models were trained on did not include semiconductor data as far as the authors of this paper know so our aim is to see how applicable these out-of-the box models are to our SEM defect classification task.

There are quite a few works on application of CNNs to semiconductor ADC. Patel et al. trained CNNs on high-resolution electron beam images with wafers patterned intentionally with defects and achieved high sensitivity and specificity [4]. Imoto et al. applied transfer learning of CNNs to classify real semiconductor fab data and concluded that their method classified defect images with high accuracy while reducing labor costs to one third [5]. Cheon et al. also applied CNNs and achieved classification accuracies of above 96% for CNNs [1]. They compared its performance to other machine learning techniques such as SAE, MP, and SVM-rbf. Phua et al. applied CNNs for defect classification of semiconductor defect SEM images that could sub-classify defects into respective sizing groups whereby defect size served as an important indicator of the root cause of the defect [6]. Compared to CNNs, there is less work published on the application of ViTs to semiconductor ADC. Qiao et al. introduced DeepSEM-Net, a novel deep learning architecture integrating CNNs with ViTs [7].

III. Proposed Methods

A. Data Collection

We first downloaded all the data from two different inspection layers from our fab system tools, cropped out annotations that would affect the training, and manually labeled and split them into training, validation, and test sets. When splitting into training, validation, and test sets, we made sure not to have images originating from the same wafer in both train and test. We collected data from two inspection layers that we call layer 1 and layer 2 in this paper. Images from layer 1 were originally 680x680 but cropped down to 340x340 to remove annotations that would affect model training. Likewise, images from layer 2 wafer cropped down from 480x480 to 340x340. Figures 2 and 3 show the SEM defects from the two inspection layers that we collected from, and Table I details the distribution of SEM defect images collected for this work. Images are then manually labeled with their defect class and resized to 224x224 to be able to be fed into the models. In total we analyzed over 7400 images and 11 different defect varieties.

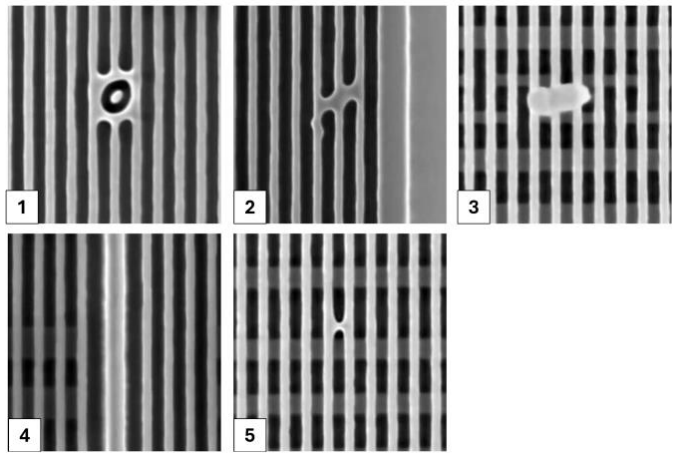

Fig. 2. SEM defect images from inspection layer 1, illustrating the six different defect types. For the purposes of this paper, defect images in this layer are labeled 1-5

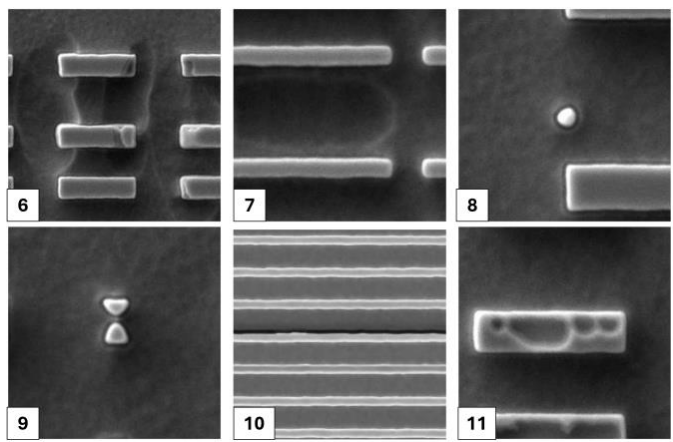

Fig. 3. SEM defect images from inspection layer 2, illustrating the five different defect types. For the purposes of this paper, defect images in this layer are labeled 6-11.

TABLE I. OVERVIEW OF LABELED SEM IMAGES COLLECTED FOR THIS STUDY. SOME DEFECTS OCCUR MORE FREQUENTLY THAN OTHERS.

| Defect Inspection Layer | Defect Class | Count |
|---|---|---|
| 1 | 1 | 288 |
| | 2 | 118 |
| | 3 | 126 |
| | 4 | 820 |
| | 5 | 842 |
| 2 | 6 | 1230 |
| | 7 | 1134 |
| | 8 | 910 |
| | 9 | 1018 |
| | 10 | 154 |
| | 11 | 796 |

## B. k-Nearest Neighbors

We used the k-nearest neighbors (k-NN) algorithm to get a preliminary performance of out-of-the-box models that haven't been trained on any of our data [8]. We first take the train images and embed them into vector space using the pretrained DinoV2 model. The vector is produced using a concatenation of the class token and flattened patch tokens from the model's final layer. When we plot these feature vectors in space, images of the same defect class form clusters. Then we also embed the test images into the vector space, compute its distance in vector space from all the labeled trained images, and then classify it based on which cluster has the most points closest to this test image's vector. This work implements weighted k-NN whereby the weight of a neighbor's "vote" is inversely proportional to its distance to the query image. The Gaussian kernel is used as the method for weighting points.

## C. t-distributed Stochastic Neighbor Embedding Plots

t-distributed stochastic neighbor embedding (t-SNE) is an algorithm for dimensionality reduction that is suitable for visualizing higher-dimensional data. It embeds points in higher-dimensional space to lower dimensions in a way that maintains similarities between points. We use this technique as an easy way to visualize how our images are separated in embedded vector space [9].

## D. Transfer Learning/Finetuning

In machine learning, transfer learning is an approach where a model that is originally trained on a set of images for a different task is applied as a starting point on a model for a separate task. The advantages of this approach are that it is known to improve model training efficiency and performance if the learned features from the original task are generic and applicable to the second task. ViT models, when pretrained on larger datasets, outperform CNNs [2]. The ViT base model, DinoV2, we use for this study has demonstrated strong applicability to secondary tasks.

To improve the results obtained from k-NN, we applied transfer learning on the DinoV2 model by finetuning it with a small amount of labeled defect SEM data. We use the AdamW optimizer, cosine scheduler with warmup, and varied learning rates. Finetuning was performed with varying amounts of training data to assess how few data we needed to still yield high classification accuracy.

## E. Semi-Supervised Learning with Pseudo-Labels

Semi-supervised learning, a popular research topic as it reduces the need for high volumes of human labeled data, simultaneously learns from both labeled and unlabeled data. It focuses on scenarios where there are large amounts of unlabeled data and limited amounts of accurately labeled data. Specifically, in this work, we implement what is known as pseudo-labeling [10]. First, unlabeled data is labeled by the supervised-trained model from our finetuning work, then we train the model using both the labeled data and these so-called pseudo-labels simultaneously. There are also other, more sophisticated, ways to implement pseudo-labels but the authors of this paper choose to first implement the most basic method to demonstrate how effective this technique already is for this use case.

## IV. Results and Discussion

In our initial experiment, k-NN, we observe that the out-of-the box model applied to this algorithm yields higher test accuracies with the more images that we provide to it. As shown in t-SNE plots of Figure 4a, the out-of-the-box model has impressive ability to separate these different defect images in embedding space before training. As shown in Figure 5, the k-NN algorithm with vectorized images via the DinoV2 model was able to achieve accuracies of over 90% with labeled images of 50 images/defect and 100 images/defect for layer 1 and layer 2 respectively. The discrepancy in the number of labeled images required could be due to the defects in layer 2 being more sophisticated than the defects in layer 1.

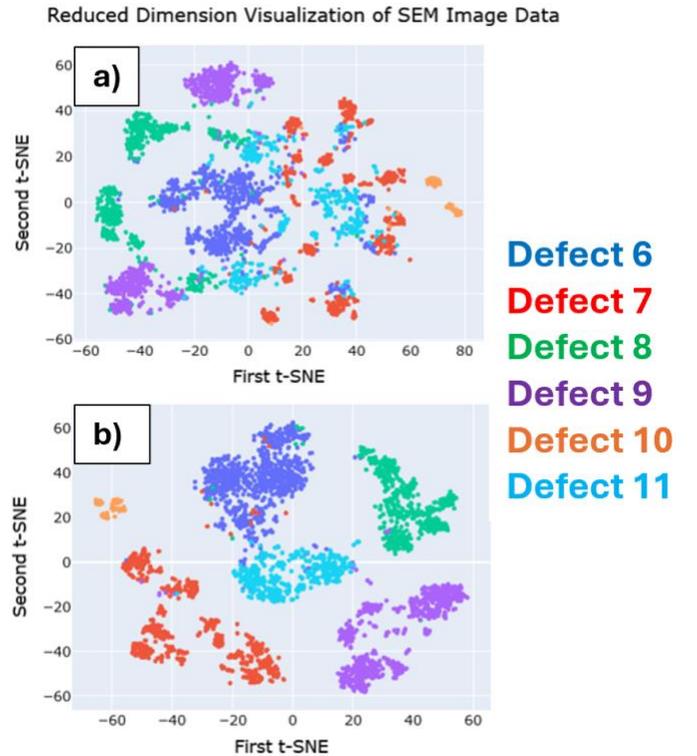

Fig. 4. t-SNE plots produced from encoded representations of SEM defect images. (a) defect images encoded by out-of-the-box model showing moderate separation of defect types. (b) defect images encoded by trained model showing better separation

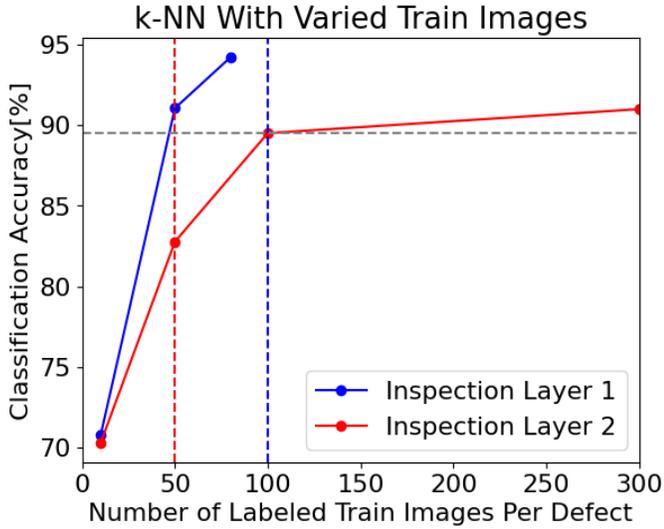

Fig. 5. Classification accuracy of k-NN algorithm using vectorized representation of SEM images embedded by out-of-the-box DinoV2 model

Next in our transfer learning/finetuning results, we can see significant improvements in test accuracy through training the model with few labeled images. As illustrated in the solid lines of Figure 6, finetuning the model was able to achieve accuracies of over 90% with 5 labeled images/defect and 15 labeled images/defect for layer 1 and layer 2 respectively. The t-SNE plot of Figure 4b shows better cluster separation between the different classes. The previous plot had more clusters for each class whereas with finetuning one can see more clearly defined clusters.

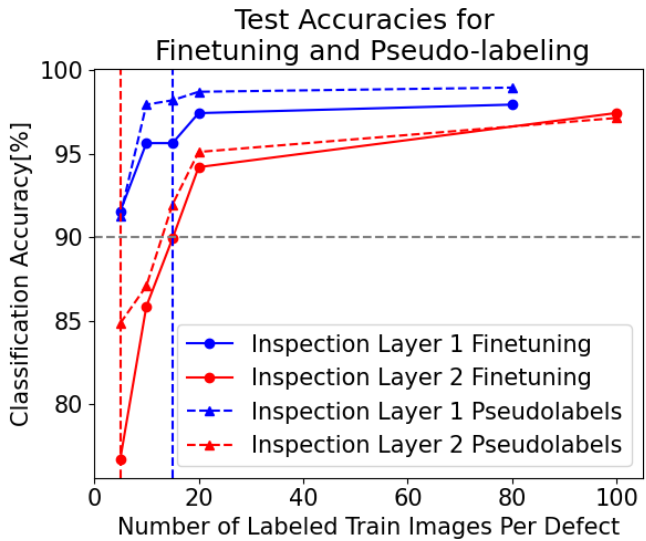

Fig. 6. Testing accuracies for models with varied numbers of train images for finetuning and pseudo-labeling methods

Finally, through pseudo-labels we were able to achieve modest improvements in accuracy. As shown in the dashed lines of Figure 6, we were able to improve accuracies by 2-6% with this implementation of pseudo-labels. Table II details the performance of the model after pseudo-label training with 15 images per defect inputted during training.

TABLE II. DETAILED RESULTS OF MODEL AFTER PSEUDO-LABELS WITH 15 IMAGES PER DEFECT USED DURING TRAINING

|  | Defect Class | Precision | Recall | F1 |
|---|---|---|---|---|
| Inspection Layer 1 | 1 | 0.94 | 1.00 | 0.97 |
|  | 2 | 0.91 | 0.91 | 0.91 |
|  | 3 | 1.00 | 0.86 | 0.93 |
|  | 4 | 0.99 | 1.00 | 0.99 |
|  | 5 | 1.00 | 0.99 | 0.99 |
| Inspection Layer 2 | 6 | 0.90 | 0.88 | 0.89 |
|  | 7 | 0.96 | 0.99 | 0.97 |
|  | 8 | 0.82 | 0.87 | 0.85 |
|  | 9 | 0.85 | 0.90 | 0.88 |
|  | 10 | 0.94 | 1.00 | 0.97 |
|  | 11 | 0.96 | 0.84 | 0.90 |

V. Conclusion

In this work we demonstrated the effectiveness of vision transformer deep learning models in classifying various types of semiconductor SEM defect images. We achieved this through transfer learning of DinoV2 with as little as 15 images per defect to achieve accuracies of above 90%. We were able to demonstrate this on two sets of real semiconductor SEM images corresponding to different inspection layers in our fab.

Results of this work can be implemented in our fab system to improve accuracy, reduce manual classification, and improve efficiency. Future directions we aim to pursue are model/algorithm improvements, scaling to more layers/defect types, deploying the model on a trusted server onsite, and explore integration of multimodal large language models for interactive interface to better assist user.